\documentclass[10pt,twocolumn,letterpaper]{article}

\usepackage[table]{xcolor}
\usepackage[pagenumbers]{iccv} 
\usepackage{makecell}
\usepackage{tabularx}
\usepackage{multirow}
\usepackage{pifont}
\definecolor{iccvblue}{rgb}{0.21,0.49,0.74}
\usepackage[pagebackref,breaklinks,colorlinks,allcolors=iccvblue]{hyperref}

\def\paperID{1073} 
\def\confName{ICCV}
\def\confYear{2025}
\definecolor{shenlan}{HTML}{a4c1f7}
\definecolor{qianlv}{HTML}{b1f49e}

\title{ Efficient Multi-Instance Generation with Janus-Pro-Dirven Prompt Parsing}
\author{
Fan Qi \textsuperscript{1}\ \ \ \ \ \  Yu Duan\textsuperscript{1} \ \ \ \ \ \     Changsheng Xu\textsuperscript{2} \\
\textsuperscript{1} Tianjin University of Technology, Tianjin, China \\
\textsuperscript{1}Institute of Automation, Chinese Academy of Sciences, Beijing, China \\
{\tt\small fanqi@email.tjut.edu.cn \ \ dyiwork@stud.tjut.edu.cn \ \ csxu@nlpr.ia.ac.cn } 
}

\begin{document}
\maketitle
\begin{abstract}
Recent advances in text-guided diffusion models have revolutionized conditional image generation, yet they struggle to synthesize complex scenes with multiple objects due to imprecise spatial grounding and limited scalability. 
We address these challenges through two key modules: 1) Janus-Pro-driven Prompt Parsing, a prompt-layout parsing module that bridges text understanding and layout generation via a compact 1B-parameter architecture, 
and 2) MIGLoRA, a parameter-efficient plug-in integrating Low-Rank Adaptation (LoRA) into UNet (SD1.5) and DiT (SD3) backbones. 
MIGLoRA is capable of preserving the base model’s parameters and ensuring plug-and-play adaptability,  minimizing architectural intrusion while enabling efficient fine-tuning.
To support a comprehensive evaluation, we create DescripBox and DescripBox-1024, benchmarks that span diverse scenes and resolutions. 
The proposed method achieves state-of-the-art performance on COCO and LVIS benchmarks while maintaining parameter efficiency, demonstrating superior layout fidelity and scalability for open-world synthesis. 
\end{abstract}    
\section{Introduction}
\label{sec:intro}

Diffusion models have achieved remarkable progress in the field of conditional image generation, especially in text-to-image applications, demonstrating significant potential through models such as GLIDE~\cite{16}, Imagen~\cite{28}, and Stable Diffusion~\cite{13}.
Multi-Instance Generation (MIG) tackles the limitations of text-to-image diffusion models in complex scene synthesis through instance-aware conditioning and enriched textual guidance.  By integrating explicit structural priors (object positions, scales) and relational semantics (interactions, occlusions), MIG bridges the gap between free-form language prompts and pixel-accurate spatial grounding, enabling coherent multi-object synthesis.

Existing MIG methods broadly follow two paradigms: \textit{ training-free approaches}~\cite{14}, which bypass fine-tuning to prioritize efficiency but suffer from attribute entanglement (e.g., fused textures/colors) in complex layouts due to inadequate instance disentanglement; and \textit{ parameter-intensive methods}~\cite{13,38}, which include the attention-based techniques~\cite{18,27} that condition on spatial coordinates yet struggle in dense layouts due to attention saturation, and the ControlNet-based frameworks~\cite{37} that add around 100M parameters ($\approx$ 78\% of UNet’s size) to enable granular instance control via parallel branches, albeit at the cost of scalability and open-world generalization. 
While these methods prioritize local spatial coherence, their reliance on parametric expansion incurs prohibitive computational overhead and compromised practical utility.
Besides, they lack mechanisms to harmonize free-form language intent with geometric precision, often failing to align global semantics (e.g., object interactions) with instance-level layout constraints.

Recent work has demonstrated that user prompt parsing~\cite{zhang2024creatilayoutsiamesemultimodaldiffusion,feng2024ranni,yang2024masteringtexttoimagediffusionrecaptioning,yu2024promptfixpromptfixphoto,park2025raretofrequentunlockingcompositionalgeneration} can significantly improve generation performance, largely attributed to the increasing accessibility of pre-trained large-scale models.
Building on this insight, we propose a dual-task MIG method that first converts the user prompt into a layout and then generates an image from that layout.
To facilitate robust layout-to-image mapping, we introduce DescripBox (2.44M), a multi-resolution dataset (512px/1024px) with two subsets: DescripBox-512 and DescripBox-1024, which ensures broad visual concept coverage and adaptability to tasks requiring varying granularity.
%
Our dual-task framework consists of two stages:
\ding{172}  Text-to-Layout, where Janus-Pro~\cite{chen2025janus} parses free-form prompts into structured layouts via a lightweight LLM adapter; and \ding{173} Layout-to-Image, where \textit{MIGLoRA} injects spatial priors into diffusion backbones (SD1.5/SD3) via mask-driven concatenation (bounding boxes → convolutional embeddings) and task-specific LoRA integration, reducing parameters by 86\% compared to SoTA~\cite{37,zhang2024creatilayoutsiamesemultimodaldiffusion}. 
A time-dependent guidance schedule balances layout adherence (early diffusion steps) and photorealism (later steps), while mask-based fusion suppresses background noise to optimize spatial fidelity. Evaluations on COCO and LVIS demonstrate that our framework achieves state-of-the-art performance with only 2M tuned parameters. 
The results are further validated on DescripBox-val, confirming scalable, multi-class, high-resolution synthesis of complex scenes with minimal computational overhead.

Our main contributions are as follows:
\begin{itemize}
    \item  We introduce a more lightweight prompt parsing module for layout generation using Janus Pro~\cite{chen2025janus} (1B parameters), which unifies understanding and generation. Our approach outperforms Qwen~\cite{qwen2.5-VL} (7B) and MiniCPM3~\cite{hu2024minicpm} (4B) in layout fidelity.  
    \item We propose \textit{MIGLoRA}, a parameter-efficient plug-in for multi-instance generation via LoRA integration across UNet (SD1.5/SD XL) and DiT (SD3) backbones, including the mask-driven feature concatenation for spatial grounding via binary masks and RoPE-Inspired Positional Encoding to boost spatial coherence by 17\%. The plug-in reduces computational costs by 40\%, supports resolutions up to 1024×1024 using task-specific LoRA ranks for efficiency.
    \item We curate DescripBox and DescripBox-1024, two benchmarks designed for rigorous evaluation of multi-instance generation across diverse scenes and resolutions.
    \item Our method achieves state-of-the-art performance on the open-ended COCO-val (5K)~\cite{lin2014microsoft} and the closed-set DescripBox-val datasets while maintaining parameter efficiency, demonstrating broad applicability and scalability.
\end{itemize}

\section{Related Work}


\subsection{Multi-Instance Text-to-Image Generation}
\label{relatedwork2}
Layout-to-image generation \cite{1}aims to synthesize realistic images that adhere to spatial layouts specified by graphical or textual input.
Early approaches, such as GAN-based models \cite{2,3,4,7}, demonstrate notable progress but are plagued by challenges such as unstable convergence \cite{5}, mode collapse \cite{6}, and limited generalization capabilities. 

Recently, diffusion models have emerged as promising alternatives that offer stable training and multimodal support for layout-based generation tasks.
Techniques such as GLIGEN~\cite{34} and ControlNet~\cite{35} directly integrate spatial constraints like bounding boxes and segmentation masks into diffusion models, improving object positioning and composition, but often require separate models for different input types, increasing system complexity. 
For example,
InstanceDiff~\cite{36} improves spatial accuracy at the instance level and the binding of attributes by incorporating multiple input forms (e.g., points, sketches, and boxes of bounding), although such multimodal input introduces additional computational overhead.
MIGC~\cite{33} employs an enhanced attention mechanism along with shadow aggregation to decompose multi-instance generation tasks into subtasks, ensuring coherence among generated objects, but complexity rises when generating numerous objects.
Conditional Attention Guidance~\cite{14} (CAG) uses a conditional attention mechanism to facilitate control over the attributes and positions of the object. However, efficiency bottlenecks still occur in complex generation scenarios. 
GLIGEN~\cite{34} provides precise management over the placement and shape of specific objects by incorporating data from the bounding box and the segmentation mask. 
This requires the development of custom models for various input types, which increases the complexity of the system.
What's more, the attention layers~\cite{1,33,34,36}, which act as implicit guidance mechanisms, require a significant number of parameters.

MtDM~\cite{11} enhances layout control by incorporating ControlNet and Adapter modules, particularly suited for intricate multi-object scenes, although the introduction of new modules significantly increases computational resource requirements.
HiCo~\cite{37} primarily supports the management of multi-object relationships in complex scenes, maintaining semantic and spatial consistency through a hierarchical attention mechanism, but incurs high computational costs when handling multiple objects. As a result, they place a substantial load on computational resources during training, limiting scalability and efficiency.
\subsection{LLM-based Prompt Parsing for Text-to-Image Generation}
LLM-based prompt parsing aims to take advantage of LLM's powerful language understanding capabilities to semantically analyze input prompts, extract key information, and generate more reasonable images. 
Recent work~\cite{lian2024llmgroundeddiffusionenhancingprompt, phung2023groundedtexttoimagesynthesisattention} has started to integrate LLMs into text-to-image diffusion frameworks. 
SLD~\cite{wu2023selfcorrectingllmcontrolleddiffusionmodels} and LayoutGPT~\cite{feng2023layoutgptcompositionalvisualplanning} utilize LLMs to decompose input prompts into multiple detailed sub-prompts and generate corresponding bounding boxes, thereby achieving reasonable layout planning. 
RPG~\cite{yang2024masteringtexttoimagediffusionrecaptioning} further leverages the chain-of-thought reasoning ability of the multimodal large language model (MLLM) to perform recaption and plan image regions, enhancing complementary regional diffusion. 
%
%
Ranni~\cite{feng2024ranni} adapts LLMs for Text-to-Panel tasks via zero-shot sequential generation (objects→attributes→layout) and fine-tuning to refine visual details like colors, enabled by structured prompts and chain-of-thought reasoning.
Createlayout~\cite{zhang2024creatilayoutsiamesemultimodaldiffusion} tames Meta-Llama-3.1-8B~\cite{grattafiori2024llama3herdmodels} into a more comprehensive and professional layout designer.
Comparing with them, our text-to-layout fine-tuning of Janus-Pro is motivated by three key advantages: 1) its unified generative-interpretative capabilities (vs. Llama’s ~\cite{touvron2023llama} understanding-only paradigm), 2) parameter-efficient adaptation (1B vs. 8B parameters) with limited training data, and 3) an iterative generate-understand-regenerate framework that refines output via dynamic user feedback.


\section{Method}
\subsection{Preliminary}

\textbf{Diffusion Backbone.}
Stable Diffusion (SD)~\cite{13} represents a state-of-the-art Latent Diffusion Model (LDM)~\cite{13} that synthesizes images through iterative denoising of Gaussian noise conditioned on text prompts. 
Architecturally, SD1.5~\cite{stable-diffusion-v1-5} and SDXL~\cite{53} utilize a UNet-driven framework with a VAE ~\cite{42} for latent encoding, a CLIP text encoder~\cite{44,23} for text-visual alignment, and a UNet denoiser for iterative noise removal, while SDXL extends this with a dual-UNet design and expanded latent space for higher resolutions. 
SD3/3.5~\cite{52} adopts a Transformer-based architecture that enhances global feature modeling and text interpretation through dual text encoders (CLIP~\cite{52} and T5~\cite{52}), Transformer-driven latent space modeling, and refined denoising strategies, resulting in improved image quality.

\subsection{Multi-instance Generation} 

\begin{figure}[ht] 
    \centering 
    \includegraphics[width=0.4\textwidth]{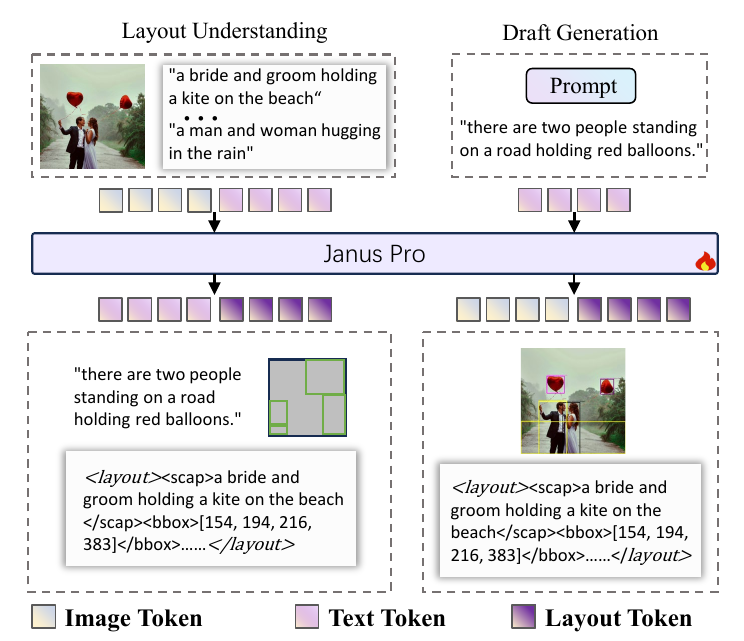} 
    \caption{Stage  $\uppercase\expandafter{\romannumeral1}$ :\textbf{Layout Understanding}, the model extracts spatial information and infers object bounding boxes from input images and textual subprompts. \textbf{Draft Generation} employs an inverse training objective to synthesize structured layouts and object descriptors, ensuring alignment between textual prompts and spatial configurations.}

    \label{fig:framework_first} 
\end{figure}

\begin{figure*}[ht] 
    \centering 
    \includegraphics[width=\textwidth]{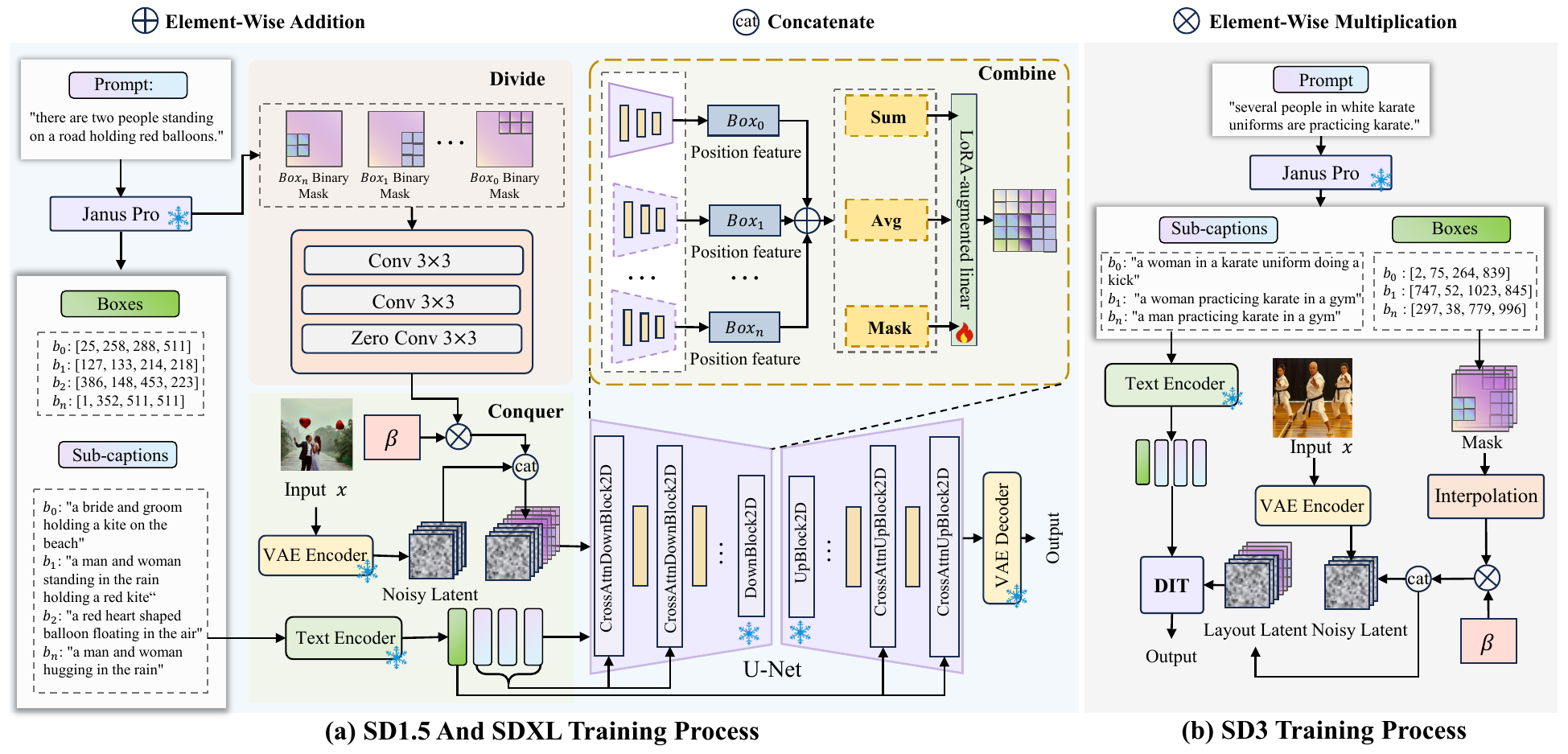} 
    \caption{
    Stage $\uppercase\expandafter{\romannumeral2}$: (a) UNet-based architecture: The bounding box encoder generates mask latents, which are concatenated with VAE-encoded image latents to form layout latents. Each layout latent is processed separately through the UNet encoder, requiring multiple passes for multiple bounding boxes. (b) DiT-based architecture: All layout latents are simultaneously fed into the DiT architecture, improving model efficiency.}
    \label{fig:framework_second} 
\end{figure*}

\paragraph{Overview} 
We decompose the entire process into two stages.
In the first stage, as shown in Fig.~\ref{fig:framework_first}, we fine-tune Janus Pro~\cite{54}, a unified model capable of both understanding and generation, by incorporating layout tokens and employing an efficient training strategy. This design enables the model to acquire planning capabilities with limited fine-tuning data and further supports the ability to understand after generation for subsequent planning.
To achieve efficient and precise multi-instance generation, we propose \textit{MIGLoRA} in the second stage, as illustrated in Fig.~\ref{fig:framework_second}.
\textit{MIGLoRA} is a LoRA plugin that works effectively across UNet-based and DiT-based models, providing high-quality multi-instance generation capabilities.  
Specifically, for UNet-based models (SD1.5/SDXL), we employ a divide-and-conquer strategy, decomposing multi-instance generation into three stages: divide (task decomposition), conquer (instance-specific feature learning via LoRA-enhanced UNet encoders), and combine (mid-layer feature fusion). 
In contrast, DiT-based SD3 leverages MM-Attention to address scalability challenges, enabling efficient training and inference while preserving spatial coherence. 
Note that we achieve fine-grained control with only 10 additional tokens, optimizing efficiency while maintaining high parameter scalability.

\paragraph{Stage  $\uppercase\expandafter{\romannumeral1}$ : Text to Layout } 
 To improve \textit{MIGLoRA}’s ability to capture prompt details and generate high-quality images, we fine-tune Janus Pro to assist in constructing coherent layouts and detailed descriptions.
 As a model that integrates both understanding and generation, Janus Pro inherently excels in layout modeling, leveraging a bidirectional synergy where generation informs understanding and understanding guides generation. 
 Based on its architecture, we employ the two-way fine-tuning process: \textit{Layout Understanding} and \textit{Draft Generation}, as illustrated in Fig.~\ref{fig:framework_first}.

To comply with the requirements of the instruction-tuned editing models~\cite{feng2024ranni,55,56}, we define a set of special tokens, \texttt{<layout></layout>}, which incorporate two sub-tokens, \texttt{<scap>} for the sub-caption and \texttt{<bbox>} for encoding layout coordinates in the format $(x_1, y_1, x_2, y_2)$.
These tokens are designed for a one-to-one correspondence between textual annotations and visual features. 
To enhance layout learning, we convert the coordinate data into a mask during training and employ bilinear interpolation to align their dimensions with the image tokens.
The resulting mask is then concatenated with the image tokens, enabling the model to effectively capture the alignment between the layout annotations and the visual features.

The \textit{Layout Understanding} phase parses input images and textual subprompts to infer object bounding boxes and spatial relationships.
For \textit{Draft Generation}, we leverage an inverse training objective: conditioned on learned object-layout correspondences, Janus Pro extracts semantic cues from global prompts to synthesize layout-consistent object descriptors.
These descriptors then guide image synthesis under spatial constraints, with bidirectional alignment refining both layout planning and object representations for semantically grounded generation.

\paragraph{Stage $\uppercase\expandafter{\romannumeral2}$: Layout to Image} 
We design divergent frameworks to accommodate architecture. 
For UNet-based models (SD 1.5, SDXL), a multi-encoder architecture is employed to enhance feature extraction and representation learning, while for DiT-based SD3, a single-encoder paradigm leverages its high-capacity Transformer backbone to optimize global coherence and generation fidelity.

Following the Divide-and-Conquer paradigm~\cite{ma2024divideconquerrethinkingtraining}, we partition multi-instance generation into three stages (Fig.~\ref{fig:framework_second}(a)):
\begin{itemize}
 \item \texttt{Divide}: Decompose the task into isolated single-instance generation subproblems.

 \item \texttt{Conquer}: Learn instance-specific latent representations via a LoRA-enhanced UNet encoder.

 \item \texttt{Combine}: Integrate features through mid-layer fusion in UNet, ensuring precise multi-instance synthesis.
 \end{itemize}

Specifically, to \texttt{Divide}, we construct a binary mask \( M_{\text{binary}} \in \{0,1\}^{H \times W} \) by setting the interiors of the box bound to 1. Three convolutional layers (channels: 16 \(\to\) 32 \(\to\) 64, stride=2) process this mask to produce coordinate embeddings:
\begin{equation}
E_{\text{bbox}} = f_{\text{conv}_3} \circ f_{\text{conv}_2} \circ f_{\text{conv}_1}(M_{\text{binary}})
\label{eq:coord_mask}
\end{equation}
where \( M_{\text{binary}} \) represents the input binary mask,  
\( f_{\text{conv}_3} \), \( f_{\text{conv}_2} \), and \( f_{\text{conv}_1} \) denote convolutional operations,  
and \( \circ \) represents the composite function.

To \texttt{Conquer,}  
CLIP encodes sub-captions \(\{\mathbf{t}_i\}\) into token embeddings, concatenated with duplicated noise latents \(\epsilon\) and layout latents \(\mathbf{z}_{\text{layout}}\). The LoRA-enhanced UNet encoder computes:
\begin{equation}
\mathbf{F}_i = \text{UNet}_{\text{LoRA}}(\epsilon \oplus \mathbf{z}_{\text{layout}} \oplus \text{CLIP}(\mathbf{t}_i))
\end{equation}

Finally, to \texttt{Combine,}  
features \(\{\mathbf{F}_i\}\) fuse via LoRA-augmented linear layers using:
1) Sum: \(\mathbf{F}_{\text{sum}} = \sum_{i=1}^n \mathbf{F}_i\), 
2) Average: \(\mathbf{F}_{\text{avg}} = \frac{1}{n}\sum_{i=1}^n \mathbf{F}_i\)  ,
3) Mask: \(\mathbf{F}_{\text{mask}} = \text{Interp} \left( \bigodot_{i=1}^{n} \left( \text{Interp}_i(M_i) \odot \mathbf{F}_i \right) \right)\), 
where \(\odot\) denotes the Hadamard product and \(\text{Interp}(\cdot)\) bilinear upsampling.

The three aforementioned fusion strategies can be flexibly utilized: Sum (element-wise addition of encoder outputs) and Average (channel-wise mean aggregation) empirically demonstrate comparable efficacy for dual-instance layouts, balancing simplicity and feature equilibrium. Mask, however, uniquely enforces structured spatial integration via bounding-box-guided bilinear interpolation, suppressing background noise while prioritizing foreground fidelity. All strategies adaptively accommodate diverse training data scenarios, enabling dynamic selection based on layout complexity and instance density.
\noindent\textbf{Single-Encoder for DiT:}  
DiT architectures in SD3/3.5 surpass SD1.5/SDXL in efficacy and efficiency via MM-Attention~\cite{52}, which optimizes multimodal fusion under scaling laws.
We augment MM-Attention with \textbf{layout tokens} (Fig.~\ref{fig:framework_second}(b)), enabling image tokens to attend to both text and layout tokens, enhancing spatial-semantic coherence.  


Bounding box coordinates are first transformed into a binary mask, which is then bilinearly interpolated into a latent layout representation \( \mathbf{Z}_{\text{layout}} \in \mathbb{R}^{10 \times 128 \times 128} \). This latent is zero-padded to create 10 slots and concatenated with noise latents. Meanwhile, sub-captions are encoded using SD3’s dual CLIP encoders, and their pooled features are concatenated with the global text tokens. This design allows for efficient parameter adaptation by adding merely 10 tokens, resulting in minimal computational overhead while maintaining excellent layout fidelity.



\subsection{Training and Inference}  

\paragraph{Training}  
We implement LoRA-enhanced spatial adaptation across SD1.5, SDXL, and SD3:  
\begin{itemize}
   
\renewcommand\labelitemi{--}
 \item  UNet (SD1.5/SDXL): Inject LoRA into the QKV and \texttt{.out} layers of the encoder, optimizing spatial attention with relative positional biases for bounding box localization. This achieves parameter-efficient adaptation while preserving model compactness.  
 \item  DiT (SD3): Extend LoRA to FFN layers via low-rank matrix decomposition, enhancing spatial-content alignment through feedforward path adaptations.  
\end{itemize}
A dynamic LoRA rank adjustment mechanism scales representation capacity with dataset size, balancing compute-accuracy tradeoffs.  

\paragraph{Inference}  




In the standard inference scheme, \(\beta\) is set to 1, which means that the entire diffusion process is influenced by the bounding-box tokens. 
Although this improves boundary alignment in generated images, it can sometimes degrade image quality. 
To mitigate this, we adopt a time-step biased sampling strategy proposed in previous work~\cite{23, 41, zhang2024creatilayoutsiamesemultimodaldiffusion}. The sampling coefficient $\beta$ varies with the diffusion step $t$:
\begin{equation}
\beta(t) = \begin{cases}
1, & t \leq 0.7*T \\
0, & t > 0.7*T
\end{cases}
\label{eq:beta_schedule}
\end{equation}
The denoising process at each step can be expressed as:
\begin{equation}
z_{t-1} = \mu_{\theta}(z_t,t) + \beta(t) \cdot g_{\phi}(z_t,b_t)
\label{eq:denoising}
\end{equation}
where $\mu_{\theta}$ represents the original denoising network and $g_{\phi}$ captures the additional bounding box information. This approach allows for a smooth transition between the bounding box-guided and standard inference stages, balancing spatial control with image quality.
The complete inference process can be written as:
\begin{equation}
p(z_{0:T} \mid c,b) = p(z_T)\prod_{t=1}^{T} p(z_{t-1} \mid z_t,c,b,\beta(t))
\label{eq:complete_inference}
\end{equation}
where $c$ represents the text condition and $b$ represents the bounding box information.In the initial stages, we determine the overall position and contours, while in the later stages, we refine high-quality details, balancing alignment accuracy with visual fidelity.

\section{Experiment}

\subsection{Dataset}  
\textbf{Training Dataset:}
In Stage $\uppercase\expandafter{\romannumeral1}$, we randomly sample 1 million instances from the GRIT-20M dataset~\cite{Kosmos2} and process them through our custom dataset filtering pipeline, resulting in a refined subset of 600K samples used for training Janus Pro.
For Stage $\uppercase\expandafter{\romannumeral2}$,
we introduce \textbf{DescripBox}, a multi-resolution dataset for layout-to-image synthesis, comprising two subsets:  
- \textit{DescripBox-512 (1.36M images)}: Aggregates and refines images from Image Aesthetic 3M~\cite{latentcat_grayscale_3M}, VQGAN pairs~\cite{41}, and UltraEdit-100k~\cite{zhao2024ultraeditinstructionbasedfinegrainedimage}, spanning landscapes, portraits, wildlife, and abstract art.  
- \textit{DescripBox-1024} (1.08M images): Extends resolution via GRIT-20M~\cite{Kosmos2}, text-to-image-2M~\cite{jackyhate_text-to-image-2M}, and DataCompDR-12M~\cite{mobileclip2024}, prioritizing complex scenes with 4+ elements.  
DescripBox-512 trains \textit{MIGLoRA(SD1.5)}, while DescripBox-1024 optimizes \textit{MIGLoRA(SD3)}, enhancing high-resolution synthesis through scale-aware spatial grounding.  

To construct \textbf{DescripBox}, we apply a systematic filtering and annotation pipeline. First, images are selected at a resolution of 512 $\times$ 512 for DescripBox-512 and 1024 $\times$ 1024 for DescripBox-1024 to ensure consistent input sizes. Next, we use RAM~\cite{45} for image tagging, Grounded-SAM~\cite{12} for bounding box and segmentation mask generation, and BLIP-V2~\cite{46} for generating descriptive prompts based on cropped regions. 
Then, images are classified by scene complexity, from simple (1-3 elements) to complex (8+ elements), ensuring a balanced distribution of scene types.
Finally, we implement a scoring system to filter out images with excessive bounding boxes, high overlap, or unclear descriptions. 

\noindent\textbf{Evaluation Dataset:}  
\textit{DescripBox-Val}:
Following a similar process, we sample 8,000 images from DescripBox-512 (average of 4.2 objects per image) and 7,000 images from DescripBox-1024 (average of 3.72 objects per image), both of which undergo automated filtering and manual verification.
\textit{Public Benchmarks}:  
We further evaluate on:  
- \textbf{COCO}~\cite{lin2014microsoft}: 5,000 validation images for multi-instance generalization.  
- \textbf{LVIS}~\cite{gupta2019lvis}: 2,800 long-tail recognition scenes to assess zero-shot spatial localization capabilities.  

\begin{table*}[t]
\centering
\resizebox{\textwidth}{!}{
\begin{tabular}{l c ccccccc ccccccc}
\toprule
\multirow{2}{*}{\Large\textbf{Method}}
&\multirow{2}{*}{\Large \textbf{Resolution}}
& \multicolumn{7}{c}{\textbf{COCO Val}}
& \multicolumn{7}{c}{\textbf{DescripBox-Val}} \\
\cmidrule(lr){3-9} \cmidrule(lr){10-16}
& & \textbf{FID$\downarrow$} & \textbf{LPIPS$\downarrow$} & \textbf{AP$\uparrow$} & \textbf{AP50$\uparrow$}  &  \textbf{AP75$\uparrow$}  & \textbf{AR$\uparrow$} &\textbf{IoU$\uparrow$}
& \textbf{FID$\downarrow$} & \textbf{LPIPS$\downarrow$} & \textbf{AP$\uparrow$} & \textbf{AP50$\uparrow$}  &  \textbf{AP75$\uparrow$}  & \textbf{AR$\uparrow$} &\textbf{IoU$\uparrow$} \\
\midrule
MtDM\cite{11}& 512& 26.8 & 0.79 & 29.0 & 36.2 & 29.3 & 36.9 & --
& 26.9 & \cellcolor[HTML]{b1f49e}0.66 & 1.63 & 28.2 & 20.0 & 7.1 & --\\
GLIGEN\cite{34}& 512& 27.1 & 0.72 & 30.3 & 40.9 & 31.7 & 40.1 & --
& 24.2 & 0.68 & 11.9 & 32.7 & 27.6 & 31.2 & -- \\
CAG\cite{14}& 512& 27.8 & 0.77 & 29.8 & 41.6 & 30.2 & 41.3 & 52.3
& 25.7 & 0.68 & 2.1 & 29.1 & 19.8 & 5.9 & 44.6 \\
MIGC\cite{33}& 512& 26.6 & 0.73 & 35.6 & 49.2 & 30.6 & 39.1 & 62.7
& 24.3 & 0.71 & 13.9 & 36.4 & 29.7 & 28.2 & 55.9  \\
HiCo\cite{37}& 512& 16.5 & 0.72 & 39.2 & \cellcolor[HTML]{b1f49e}58.1 & 40.1 & 48.6 & --
& 15.1 & 0.73 &\cellcolor[HTML]{b1f49e}20.0 & 38.9 & \cellcolor[HTML]{a4c1f7}32.1 &  36.9 & --  \\
InstDiff\cite{36}& 512& 23.9 & 0.73 & 38.8 & 55.4 & 38.6 & \cellcolor[HTML]{b1f49e}52.9 & 63.9
& 16.9 & 0.71 & 14.8 & 37.1 & 29.8 & 36.5 &61.6 \\
\textit{MIGLoRA(SD1.5)}
&512&\cellcolor[HTML]{b1f49e}16.0&\cellcolor[HTML]{b1f49e}0.71&\cellcolor[HTML]{b1f49e}39.5&57.8&\cellcolor[HTML]{b1f49e}40.1&52.1&\cellcolor[HTML]{b1f49e}64.0& \cellcolor[HTML]{b1f49e}14.7&  0.71& 15.1&\cellcolor[HTML]{b1f49e}39.2& 30.0& \cellcolor[HTML]{b1f49e}37.0&  \cellcolor[HTML]{b1f49e}61.9\\
\textit{MIGLoRA\textsuperscript{JP}(SD1.5)}& 512& \cellcolor[HTML]{a4c1f7}15.7 & \cellcolor[HTML]{a4c1f7}0.65 & \cellcolor[HTML]{a4c1f7}40.1 & \cellcolor[HTML]{a4c1f7}58.3 & \cellcolor[HTML]{a4c1f7}40.2 &  \cellcolor[HTML]{a4c1f7}53.6 & \cellcolor[HTML]{a4c1f7}64.5
& \cellcolor[HTML]{a4c1f7}14.3 &  \cellcolor[HTML]{a4c1f7}0.57 &  \cellcolor[HTML]{a4c1f7}23.6 &  \cellcolor[HTML]{a4c1f7}39.6 &  \cellcolor[HTML]{b1f49e}30.1 & \cellcolor[HTML]{a4c1f7}37.6 & \cellcolor[HTML]{a4c1f7}62.0 \\
\midrule
\textit{MIGLoRA\textsuperscript{JP}(SDXL)}& 768&15.7&0.68& 39.7&58.2&40.2&53.8&65.1 &14.5&0.55&24.5&40.1&30.3&38.0 &62.6\\
\midrule
CreatiLayout\cite{zhang2024creatilayoutsiamesemultimodaldiffusion}& 1024&20.1&0.78&38.5&55.1&38.6&53.0&64.1 &16.5&0.70&23.1&38.4&28.9&36.3&61.5\\
\textit{MIGLoRA\textsuperscript{JP}(SD3)}& 1024&\cellcolor[HTML]{a4c1f7}14.9&\cellcolor[HTML]{a4c1f7}0.65&\cellcolor[HTML]{a4c1f7}40.2&\cellcolor[HTML]{a4c1f7}59.3&\cellcolor[HTML]{a4c1f7}40.2&\cellcolor[HTML]{a4c1f7}54.1&\cellcolor[HTML]{a4c1f7}65.3 &\cellcolor[HTML]{a4c1f7}14.1&\cellcolor[HTML]{a4c1f7}0.56&\cellcolor[HTML]{a4c1f7}25.1&\cellcolor[HTML]{a4c1f7}40.6&\cellcolor[HTML]{a4c1f7}30.4&\cellcolor[HTML]{a4c1f7}38.3&\cellcolor[HTML]{a4c1f7}64.2\\
\bottomrule
\end{tabular}
}
\caption{
Quantitative comparison of \textit{MIGLoRA} (with/without Janus-Pro-Powered Prompt Parsing, denoted as the superscript with JP) against state-of-the-art baselines on COCO-Val and DescripBox-Val for text-to-image generation. The best and second-best results are highlighted in blue and green, respectively.
}
\label{tab:1}
\end{table*}

\subsection{Experimental Settings}

\textbf{Text to Layout}: We set the batch size to 128 and perform 94K iterations on the A800 GPU to ensure sufficient model training.
\textbf{Layout to Image}:
\textit{MIGLoRA} employs LoRA~\cite{49} fine-tuning across diffusion backbones with the following configurations: 

\begin{itemize}[leftmargin=12pt]  
  \renewcommand\labelitemi{$\blacksquare$}
    \item \textbf{SD1.5 \& SDXL}: AdamW~\cite{51} optimizer, learning rate $10^{-4}$, batch size 256.  
   
     \textit{MIGLoRA\textsuperscript{JP}(SD1.5)}: 40K iterations, LoRA rank 256.  
       
     \textit{MIGLoRA\textsuperscript{JP}(SDXL)}: 65K iterations, LoRA rank 256.  

    \item \textbf{SD3}: AdamW optimizer, learning rate $10^{-4}$, batch size 256, trained on 16 H100 GPUs.  
    \begin{itemize}[leftmargin=20pt]  
        \item \textit{MIGLoRA\textsuperscript{JP}(SD3)}: 70K iterations, LoRA rank 256 (optimized via rank scaling for Transformer efficiency).  
    \end{itemize}  
\end{itemize}  

\begin{figure*}[t]
\centering
\includegraphics[width=0.9\textwidth]{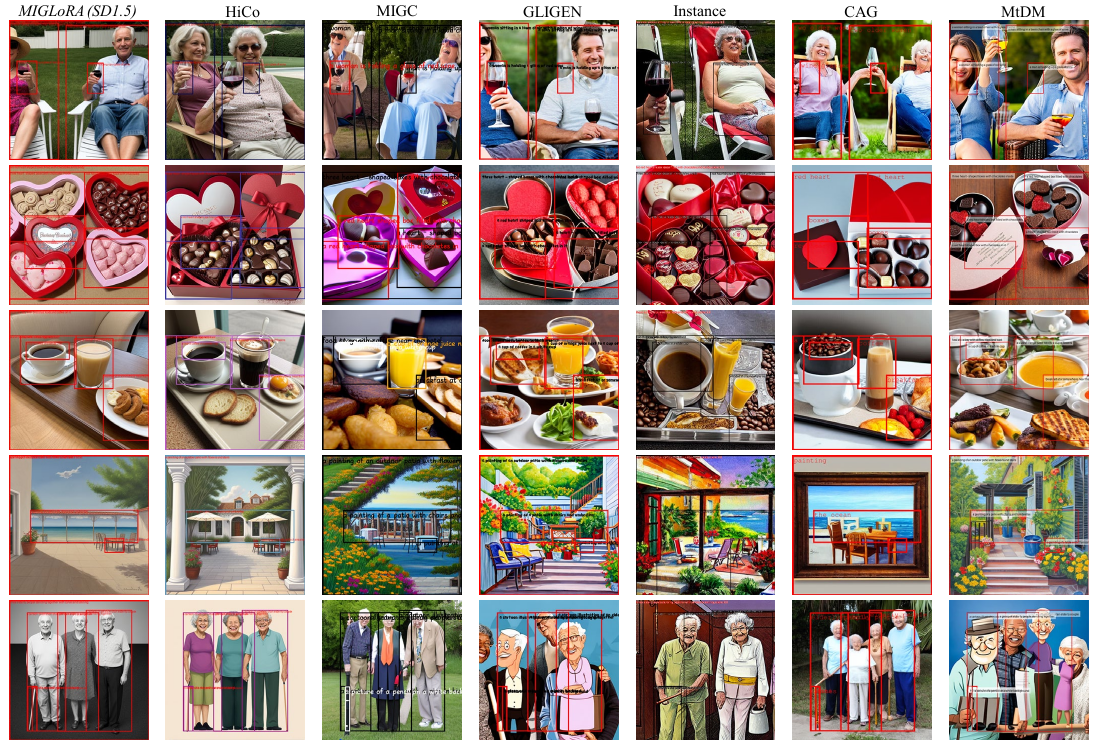} 
\caption{Qualitative comparison with SOTA methods on COCO val 512×512. Compared to baseline methods (CAG ~\cite{14}, MtDM~\cite{11}, MIGC~\cite{33}, InstanceDiff ~\cite{36}, GLIGEN ~\cite{34}, and HiCo ~\cite{37}), \textit{MIGLoRA(SD1.5)} demonstrates superior performance in composing multiple independent concepts (\(\geq 4\) objects) while maintaining better spatial relationships and visual quality.}
\label{fig3:val1.5}
\vspace{-1.0em}
\end{figure*}

\subsection{Evaluation Metrics \& Baselines}
\textbf{Evaluation Metrics:}
To evaluate performance on COCO and DescripBox-Val datasets, we use the following metrics: FID, LPIPS, AP, AP$_{50}$, AP$_{75}$, AR, and IoU. Specifically, lower FID and LPIPS values indicate better performance, while higher values are better for the other metrics.
\textbf{Metrics on LVIS:}
CLIP$_{\text{local}}$ and IoU\textsubscript{local} are evaluation metrics specifically designed for zero-shot learning tasks. Traditional AP metrics rely on label distributions and class information, which may not be available in zero-shot tasks. 
\textbf{Baselines:}
We compare our method with several SOTA MIGC methods: MtDM~\cite{11}, GLIGEN~\cite{34},
CAG~\cite{14}, MIGC~\cite{33}, HiCo~\cite{37}, InstanceDiff~\cite{36}, and CreatiLayout~\cite{zhang2024creatilayoutsiamesemultimodaldiffusion}, as detailed in \cref{relatedwork2}.


\subsection{Quantitative Results}
\textbf{COCO}: As shown in Table~\ref{tab:1}, \textit{MIGLoRA(SD1.5)} reduces FID from 16.5 to 16.0 and LPIPS from 0.72 to 0.71, significantly improving image quality. Meanwhile, AP improves from 39.2 to 39.5, and IoU grows from 63.9 to 64.0. \textit{MIGLoRA\textsuperscript{JP}(SD1.5)} has achieved comprehensive improvements over \textit{MIGLoRA(SD1.5)} in all metrics, demonstrating its strong capability in layout control. As for \textit{MIGLoRA\textsuperscript{JP}(SD3)}, it outperforms CreatiLayout in all metrics. This demonstrates that our method can effectively balance image quality and layout accuracy while maintaining higher computational efficiency.

\textbf{DescripBox-Val}: As shown in Table~\ref{tab:1}, \textit{MIGLoRA(SD1.5)} has made some progress in FID, AR, and IoU. \textit{MIGLoRA\textsuperscript{JP}(SD1.5)} significantly boosts performance across multiple metrics. It achieves the lowest FID of 14.3 and the highest AR score of 37.6, significantly improving the quality of images. For \textit{MIGLoRA\textsuperscript{JP}(SD3)}, our model still outperforms CreatiLayout in all metrics, showcasing its strong capability. 

\begin{table}[t]
\center
\footnotesize
\setlength{\tabcolsep}{0.6mm}
\renewcommand{\arraystretch}{1.0}
\begin{tabular}{c|ccccccc}
\textbf{Methods} &MtDM  & GLIGEN  &CAG  &MIGC &HiCo &InstDiff &\textit{MIGLoRA}\\ \midrule[1pt]
\textbf{CLIP$_{\text{local}}$$\uparrow$} &20.11  &21.01  &19.86  &22.03 &22.57 &\cellcolor[HTML]{b1f49e}22.41 &\cellcolor[HTML]{a4c1f7}22.61\\
\textbf{IoU\textsubscript{local}$\uparrow$} &22.01  &38.27  &20.10 &42.62 &42.86 &\cellcolor[HTML]{b1f49e}44.50 &\cellcolor[HTML]{a4c1f7}45.10 \\ 
\end{tabular}
\caption{Quantitative comparisons of zero-shot spatial localization capabilities between our method and SOTA on the LVIS~\cite{gupta2019lvis} dataset.
}
\vspace{-3mm}
\label{tab:2}
\end{table}

\textbf{LVIS}: As shown in Table~\ref{tab:2}, 
we evaluate the zero-shot performance of our model on the LVIS dataset, demonstrating moderate improvements with fewer parameters, avoiding expensive full-dataset training or complex attention mechanisms.
Compared to HiCo~\cite{5}, based on ControlNet~\cite{35}, our method achieves higher CLIP$_{\text{local}}$ and IoU\textsubscript{local} scores with fewer parameters, providing competitive performance and efficiency. 

\subsection{Qualitative analysis}
\begin{figure}[ht]
\centering
\includegraphics[width=0.4\textwidth]{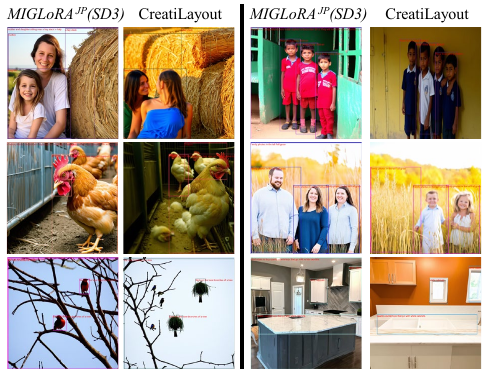} 
\caption{Qualitative comparison with SOTA method on DescripBox-Val. Compared to CreatiLayout~\cite{zhang2024creatilayoutsiamesemultimodaldiffusion}, our model uses fine-tuning of Stable Diffusion 3 to generate high-quality 1024 $\times$ 1024 images in the task of layout-based image generation.}
\label{fig4:val3}
\vspace{-3mm}
\end{figure}

\begin{figure}[ht]
\centering
\includegraphics[width=0.4\textwidth]{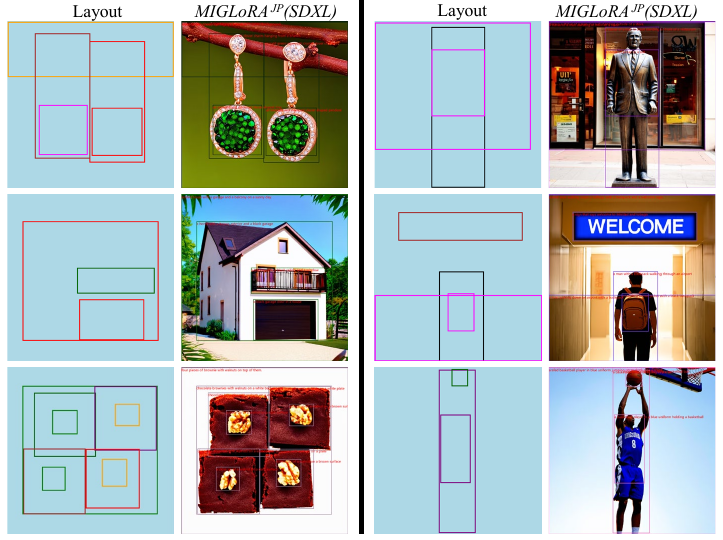} 
\caption{The experimental results of \textit{MIGLoRA\textsuperscript{JP}(SDXL)} show that our model can generate satisfactory images in complex scenarios.}
\label{fig5:valxl}
\vspace{-3mm}
\end{figure}

We qualitatively analyze the model's performance in spatial and textual consistency to evaluate its ability to generate images that adhere to spatial requirements while matching the textual descriptions. Figure \ref{fig3:val1.5}, \ref{fig4:val3}, and \ref{fig5:valxl} show a comparison of the consistency of generated images under different conditions. Our method excels in both spatial and textual consistency. For example, in the task of generating character images, our method accurately captures layout information, such as an image of two elderly people sitting together with wine glasses. In the food image generation task, only our method precisely locates the position of each food item on the plate and matches the corresponding textual description, generating a reasonable plate layout. Additionally, our method significantly outperforms other baseline models in the fusion of elements. Our method shows significant improvement in processing long captions. The experimental results for long captions and additional visualizations are provided in the supplementary materials.

\subsection{Ablation Studies}

To validate the effectiveness of each component in our method, we conduct systematic ablation studies. Table \ref{tab:3} presents the experimental results, focusing on three key steps: Janus-Pro-driven Prompt Parsing (JP), Divide (Div), and Combine (Comb).

\textbf{Impact of Divide}:
Comparing rows 1 and 2, the addition of our critical Divide operation shows significant improvements across all metrics. With this essential component, the FID score drops further to 19.66, while the CLIP\textsubscript{local} increases from 9.61 to 18.27. 
This demonstrates that Divide is crucial for spatial understanding and layout control. 

\textbf{Exploration of Combine}:
Through our ablation studies examining feature fusion methods while maintaining the Divide stage, we explore three ways: summation (sum), averaging (avg), and mask-based fusion (mask). The results show that the mask-based fusion method achieves the best performance across all evaluation metrics. Specifically, it achieves the lowest FID score of 15.70, along with significant improvements in AR (53.60), AP (40.10), and CLIP$_{\text{local}}$ (22.61) compared to alternative fusion methods. 

\textbf{Assessment of Janus-Pro-driven Prompt Parsing}:
Table~\ref{tab:llm} presents our analysis of LLMs' layout accuracy and the quality of images generated from these layouts. Accuracy evaluates the precision of the generated bounding boxes, ensuring correct coordinates within image boundaries. The results show that Janus-Pro-1B achieves excellent performance with fewer parameters: its Acc score reaches 90.65, comparable to the larger Qwen2.5-VL-7B~\cite{qwen2.5-VL} and outperforming MiniCPM3-4B~\cite{hu2024minicpm}. Furthermore, Janus-Pro-1B excels on the ReAcc metric with a score of 93.90, highlighting its powerful capabilities in layout planning and image understanding. Based on Table ~\ref{tab:3}, when Janus Pro is applied (rows 3), the FID score decreases to 16.21, while the other metrics show a reduction compared to the previous stage(row 2). These results demonstrate that Janus-Pro-Driven Prompt Parsing plays a key role in enhancing model performance.

\textbf{Impact of LoRA Rank:} 
We compare the impact of different LoRA rank values on our model under different architectures. As shown in Table~\ref{tab:4}, for \textit{MIGLoRA\textsuperscript{JP}(SD1.5)}, the FID decreases as the rank value rises, from 16.0 to 14.3. Similarly, for \textit{MIGLoRA\textsuperscript{JP}(SD3)}, the FID score continuously decreases with increasing rank value, reaching 14.1. This suggests that higher LoRA rank values lead to improved model performance. Additionally, we randomly sample 1K data as the training set, set the LoRA rank to 8, and find that our model remains effective under these conditions.
The results in Table~\ref{tab:4} also validate the scaling law, further suggesting that increasing the parameter size of SD3 without modifying the LoRA rank—such as extending to SD3.5 (8B) or FLUX.dev (12B)—can potentially lead to even better performance.

\begin{table}[t]
\center
\fontsize{8}{8}\selectfont 
\setlength{\tabcolsep}{5pt} 
\renewcommand{\arraystretch}{1.5}
\begin{tabular}{c|ccc}
\textbf{Layout Planning} &Acc$\uparrow$ & Quality$\uparrow$ & ReAcc$\uparrow$\\ 
\midrule[1pt]
\textbf{MiniCPM3-4B~\cite{hu2024minicpm}} &77.25  &60.11  &75.34\\
\textbf{Qwen2.5-VL-7B~\cite{qwen2.5-VL}} &\cellcolor[HTML]{a4c1f7}92.58  &\cellcolor[HTML]{a4c1f7}89.22  &\cellcolor[HTML]{b1f49e}83.21\\ 
\textbf{Janus-Pro-1B~\cite{54}} &\cellcolor[HTML]{b1f49e}90.65  &\cellcolor[HTML]{b1f49e}72.10  &\cellcolor[HTML]{a4c1f7}93.90\\ 
\bottomrule
\end{tabular}
\caption{Quantitative evaluation of image quality and accuracy of object positions in layout analysis using different LLMs. ReAcc measures the accuracy of LLMs in correcting erroneous layouts when these layouts are resubmitted to the LLMs.}
\label{tab:llm}
\end{table}

\begin{table}[t]
\centering
\fontsize{8}{8}\selectfont 
\setlength{\tabcolsep}{5pt} 
\renewcommand{\arraystretch}{1.5}
\begin{tabular}{ccc|cccc} 
\textbf{JP}&\textbf{Div}& \textbf{Comb} & \textbf{FID$\downarrow$} & \textbf{AR$\uparrow$} & \textbf{AP$\uparrow$} & \textbf{CLIP$_{\text{local}}$$\uparrow$}\\
\midrule[1pt]
$\times$&$\times$  & sum & 28.92 & 16.51 & 5.23 & 9.61 \\
$\times$&$\checkmark$  & sum & 19.66 & 29.22 & 10.32 & 18.27\\
$\checkmark$&$\checkmark$ & sum & \cellcolor[HTML]{b1f49e}16.21 & \cellcolor[HTML]{b1f49e}32.63 & \cellcolor[HTML]{b1f49e}12.36 & \cellcolor[HTML]{b1f49e}20.19  \\
$\checkmark$&$\checkmark$  & avg & 16.35 & 31.11 & 12.14 & 19.98 \\
$\checkmark$&$\checkmark$  & mask & \cellcolor[HTML]{a4c1f7}15.70 & \cellcolor[HTML]{a4c1f7}53.60 & \cellcolor[HTML]{a4c1f7}40.10 & \cellcolor[HTML]{a4c1f7}22.61 \\
\bottomrule[1pt]
\end{tabular}
\caption{Our ablation study results. We systematically analyze the impact of three key components by removing or replacing them. 
}
\label{tab:3}
\vspace{-5mm}
\end{table}

\begin{table}[t]
\vspace{4mm}
\center
\fontsize{8}{8}\selectfont 
\setlength{\tabcolsep}{3mm}
\renewcommand{\arraystretch}{1.0}
\begin{tabular}{c ccc ccc}
\toprule
\multirow{2}{*}{\textbf{Rank}}
&\multicolumn{3}{c}{\textit{MIGLoRA\textsuperscript{JP}(SD1.5)}}
&\multicolumn{3}{c}{\textit{MIGLoRA\textsuperscript{JP}(SD3)}} \\
\cmidrule(lr){2-4} \cmidrule(lr){5-7} 
&64&128&256&64&128&256\\ 
\midrule
\textbf{FID$\downarrow$} & 16.0 & 15.5 & \cellcolor[HTML]{a4c1f7}14.3& 15.9 &14.5 & \cellcolor[HTML]{a4c1f7}14.1 \\
\bottomrule[1pt]
\end{tabular}
\caption{Ablation study results comparing the model's FID scores under different LoRA Rank configurations, illustrating performance variations across each rank setting.}
\vspace{-4mm}
\label{tab:4}
\end{table}



\textbf{Why Not Janus Pro Alone?}  
Janus Pro has critical constraints for layout-to-image synthesis:  
1. \textbf{Resolution Bottleneck}: It generates images at \(384\!\times\!384\) resolution, which is insufficient for high-fidelity generation (\(1024\!\times\!1024\) in our framework), limiting detail preservation in complex scenes.  
2. \textbf{Attention Collapse}: As the mask count exceeds 4, self-attention layers suffer quadratic token growth (\(N_{\text{tokens}} \propto N_{\text{masks}}^2\)), weakening interactions between text and bounding box tokens.

\section{Conclusions \& Furture Work}
Our method for MIG achieves significant improvement in handling multiple boxes and long captions in various test sets.
Through experiments, our model can achieve superior performance with fewer parameters, which highlights our core advantage of delivering stronger generation performance with less complexity.
In the future, we plan to further evaluate the performance of this method on SD3.5, FLUX, and other larger-scale models.
And we will investigate the seamless integration of the two-stage generation paradigm into an end-to-end synthesis framework.

{
    \small
    \bibliographystyle{ieeenat_fullname}
    \bibliography{main}
}

\end{document}